\documentclass{ecai} 
\usepackage{latexsym}
\usepackage{amssymb}
\usepackage{amsmath}
\usepackage{amsthm}
\usepackage{booktabs}
\usepackage{enumitem}
\usepackage{graphicx}
\usepackage{color}

\newcommand{\rom}[1]{\uppercase\expandafter{\romannumeral #1\relax}}

\newcommand{\BibTeX}{B\kern-.05em{\sc i\kern-.025em b}\kern-.08em\TeX}

\begin{document}

%%%%%%%%%%%%%%%%%%%%%%%%%%%%%%%%%%%%%%%%%%%%%%%%%%%%%%%%%%%%%%%%%%%%%%%%

\begin{frontmatter}

%%% Use this command to specify your submission number.
%%% In doubleblind mode, it will be printed on the first page.

\paperid{123} 

%%% Use this command to specify the title of your paper.

\title{Holistic analysis on the  sustainability of Federated Learning lifecycle in real-world industrial settings}

%%% Use this combinations of commands to specify all authors of your 
%%% paper. Use \fnms{} and \snm{} to indicate everyone's first names 
%%% and surname. This will help the publisher with indexing the 
%%% proceedings. Please use a reasonable approximation in case your 
%%% name does not neatly split into "first names" and "surname".
%%% Specifying your ORCID digital identifier is optional. 
%%% Use the \thanks{} command to indicate one or more corresponding 
%%% authors and their email address(es). If so desired, you can specify
%%% author contributions using the \footnote{} command.

\author[A]{\fnms{Hongliu}~\snm{CAO}\orcid{0000-0002-1326-8159}\thanks{Corresponding Author. Email: hongliu.cao@amadeus.com.}}
%\author[B]{\fnms{Second}~\snm{Author}\orcid{....-....-....-....}\footnotemark}
%\author[B,C]{\fnms{Third}~\snm{Author}\orcid{....-....-....-....}} 

\address[A]{Amadeus SAS}
%\address[B]{Short Affiliation of Second Author and Third Author}
%\address[C]{Short Alternate Affiliation of Third Author}

%%% Use this environment to include an abstract of your paper.

\begin{abstract}
In light of emerging legal requirements and policies focused on privacy protection, there is a growing trend of companies across various industries adopting Federated Learning (FL).
This decentralized approach involves multiple clients or silos, collaboratively training a global model under the coordination of a central server while utilizing their private local data. Unlike traditional methods that require data sharing and transmission, Cross-Silo FL allows clients to share model updates rather than raw data, thereby enhancing privacy. Despite its growing adoption, the carbon impact associated with Cross-Silo FL remains poorly understood due to the limited research in this area. Furthermore, most existing studies rely on simulated environments rather than real-world scenarios.
This study seeks to bridge this gap by evaluating the sustainability of Cross-Silo FL throughout the entire AI product lifecycle in real-world industrial settings, extending the analysis beyond the model training phase alone. We systematically compare this decentralized method with traditional centralized approaches and present a robust quantitative framework for assessing the costs and CO2 emissions in real-world Cross-Silo FL environments. Our findings indicate that the energy consumption and costs of model training are comparable between Cross-Silo Federated Learning and Centralized Learning. However, the additional data transfer and storage requirements inherent in Centralized Learning can result in significant, often overlooked CO2 emissions. Moreover, we introduce an innovative data and application management system that integrates Cross-Silo FL and analytics, aimed at improving the sustainability and economic efficiency of IT enterprises. This study highlights the real-world applicability and benefits of Federated Learning, providing valuable insights and lessons learned that can be leveraged to enhance the deployment of AI methods in industrial applications.
\end{abstract}

\end{frontmatter}

\section{Introduction}

The Information and Communication Technology (ICT) sector has experienced substantial and rapid growth over the past few decades, contributing significantly to global CO2 emissions \cite{belkhir2018assessing}. This upward trajectory is expected to accelerate, fueled by the increasing reliance on technology such as Artificial Intelligence (AI) in everyday activities \cite{belkhir2018assessing,freitag2021real, cao2024recent}. Innovations in Machine Learning (ML) and Deep Learning (DL) have greatly enhanced efficiency and accuracy across various sectors \cite{cao2019random, cao2018improve,cao2025enhancing}. However, these advancements have also resulted in the development of more complex models with larger parameter sizes and larger training data size, necessitating considerable energy consumption and substantial water usage for cooling data centers \cite{bolon2024review}.
For example, the training of the GPT-3 model which comprises 175 billion parameters  reportedly consumed around 1,287 MWh of electricity, leading to carbon emissions estimated at approximately 500 metric tons of CO2 \cite{ajay,cho2023,faiz2023llmcarbon}. To contextualize this, the energy used is sufficient to power about 121 average households in the United States for a full year \cite{bolon2024review}. Beyond the training phase, other components of the AI product lifecycle, including data storage, model inference, and model management, also have substantial environmental impacts. Consequently, it is imperative to conduct comprehensive studies on the carbon footprint associated with the entire lifecycle of AI products. Furthermore, the development and implementation of effective strategies to enhance sustainability within the ICT sector are essential.

In traditional AI model training, data is usually aggregated in a centralized manner, a process known as Centralized Learning (CL). However, CL presents significant vulnerabilities as centralized data repositories are prone to cyber-attacks, potentially leading to grave repercussions. Furthermore, when the data encompass personal or proprietary information, it also raises profound privacy concerns and complicates the collection and training processes within centralized systems \cite{qi2024model}. 
To mitigate these privacy issues, Federated Learning (FL) \cite{konevcny2016federated} has emerged as an innovative solution. FL is a distributed learning framework that emphasizes privacy preservation by allowing multiple clients to collaboratively train a global model under the guidance of a central server \cite{chen2024federated, cao2023multi}. Instead of sharing raw data, clients utilize their private local datasets to train models and subsequently share the model updates. When the clients participating in the training are small or distributed entities with limited local data, the process is known as cross-device Federated Learning. On the other hand, Cross-Silo Federated Learning involves a smaller number of clients, typically organizations or companies, each possessing substantial local datasets \cite{li2022federated}.

Cross-Silo Federated Learning is increasingly being adopted on a global scale by numerous companies to adhere to emerging legal requirements and privacy protection policies \cite{qiu2023first}. Despite its growing implementation, there is a notable lack of comprehensive understanding regarding its environmental impact. This research aims to fill this gap by examining the sustainability of Cross-Silo FL across the entire lifecycle of AI products, rather than limiting the focus to the model training phase. To achieve this, we perform a systematic comparison between decentralized Cross-Silo FL and traditional centralized methodologies, employing a rigorous quantitative framework to evaluate both the financial costs and CO2 emissions associated with real-world Cross-Silo FL scenarios. Additionally, the study explores potential strategies for companies to reduce their carbon footprint through the adoption of Cross-Silo FL, an area that has not been extensively studied. The key contributions of this research are as follows:
\begin{enumerate}
    \item 	A novel framework for the analysis of energy and carbon footprints of Cross-Silo FL across the AI product life cycle (including data storage, data transmission, model training, communication, etc.) instead of focusing only on the model training in the literature, with the comparison to the centralized method.
   
    \item 	Real-world FL environments: this study reports and analyzes the cost and carbon footprint of Cross-Silo Federated Learning in multiple industrial scenarios. These experiments are conducted using actual hardware, network, and security settings of Federated Learning on the Cloud, as opposed to simulated environments.
    
    \item A novel system based on Cross-Silo Federated Learning and Federated Analytics is introduced, with the aim to improve the sustainability, minimize data, model, and application redundancy within organizations while simultaneously enhancing data privacy.
\end{enumerate}

The paper is organized as follows: the background and related works are introduced in Section 2. In Section 3, the novel framework for the analysis of energy and carbon footprints in Cross-Silo FL is introduced. We describe the experiments comparing the CO2 emission and cost of Cross-Silo Federated Learning with Centralized Learning in Section 4 and the proposed sustainable data and application management system in Section 5. Finally, the conclusion and future works are given in Section 6.

\section{Related Works}

\begin{figure}
	    \centering
	    \includegraphics[width=0.5\textwidth]{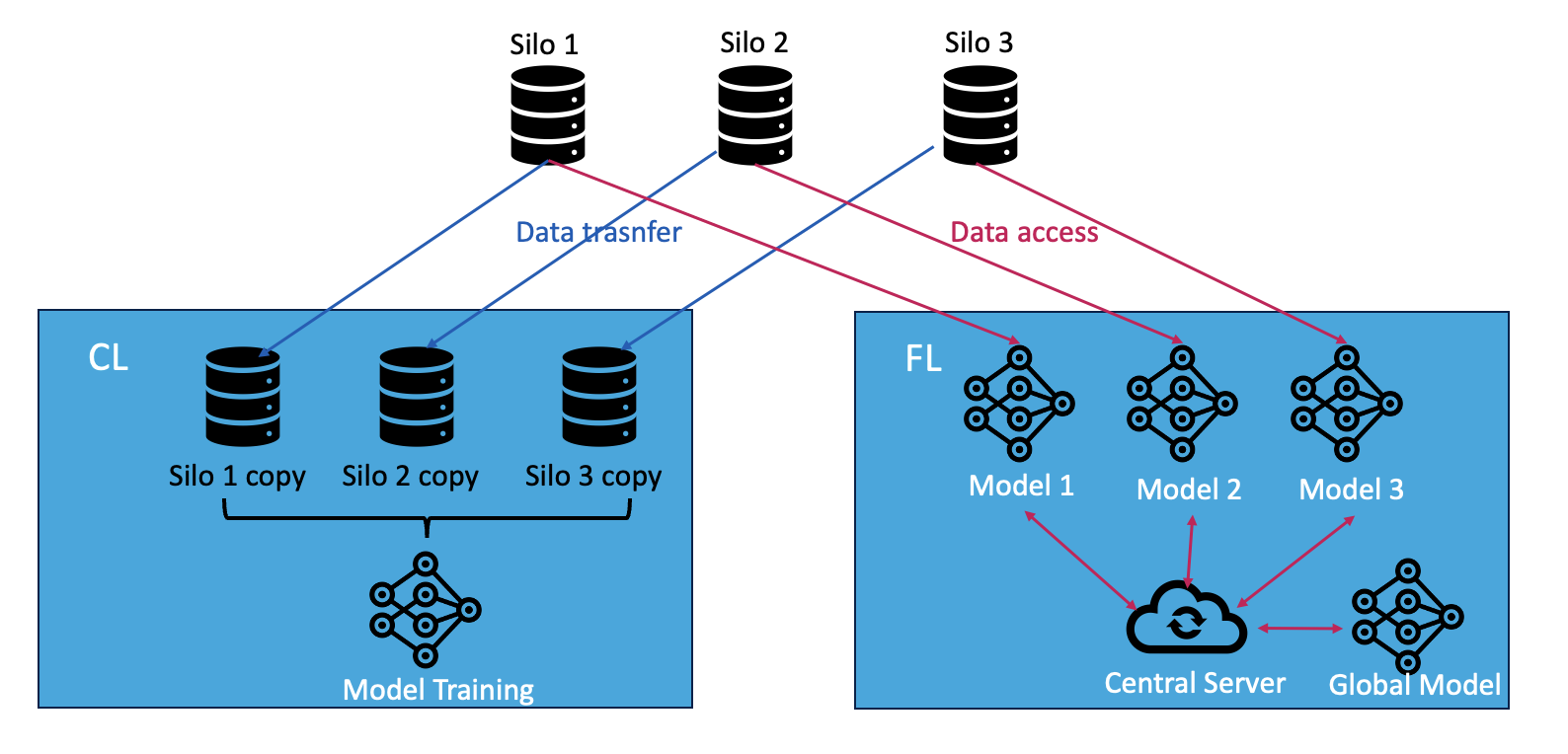}
	    \caption{The difference between Federated Learning (FL) and Centralized Learning (CL).}
	    \label{fig:clfl}
	\end{figure}

Despite fast development and impressive results of ML and DL, the data privacy and environmental concerns induced by the training and inference of these models have also been raised in recent years. 
In response, alternatives to Centralized Learning, such as Federated Learning and Federated Analytics, have been developed. Federated Learning facilitates collaborative learning in a distributed manner under the coordination of a central server, as illustrated in Figure \ref{fig:clfl}. The concept of Federated Analytics was introduced by Google in 2020, defined as “collaborative data science without data collection, utilizing the Federated Learning infrastructure but excluding the learning component” \cite{fa, elkordy2023federated}. Both Federated Learning and Federated Analytics enhance data privacy by eliminating the need for data aggregation for learning or analysis purposes.

Numerous studies emphasize the privacy benefits of Federated Learning and Federated Analytics. Beyond privacy, these methodologies also present opportunities for enhanced sustainability compared to Centralized Learning, particularly by reducing the need for extensive data transfers in the era of big data. As demonstrated in Figure \ref{fig:clfl}, federated approaches access data directly rather than duplicating it, which can significantly cut down on data transfer across networks. This reduction in data redundancy not only mitigates the carbon footprint associated with additional data storage but also has the potential to lower the energy consumption of data centers. Furthermore, it is crucial to examine the environmental implications of training models in Centralized Learning versus Federated Learning, as their computational and memory demands differ markedly.

Most recent studies on Federated Learning primarily address issues related to  Non-Independent and Identically Distributed (Non-IID) data and trustworthy Federated Learning. There is only a limited amount of research focusing on the sustainability aspects of Federated Learning.
For instance, Savazzi et al \cite{savazzi2022energy} initialized an energy and carbon footprint analysis of distributed and Federated Learning. They quantify both the energy footprints and the carbon equivalent emissions for vanilla FL methods and consensus-based fully decentralized approaches. Case studies from the 5G industry show that sustainable distributed learning depends on communication efficiency, learner population size, and a trade-off between energy consumption and test accuracy \cite{savazzi2022energy}.
Similarly in the work of \cite{qiu2023first}, the authors offered the first-ever systematic study of the carbon footprint of Federated Learning with a rigorous model to quantify the carbon footprint, facilitating the investigation of the relationship between FL design and carbon emissions.  Their findings reveal that FL can emit significantly more carbon than centralized training, but in some cases, it can be comparable due to lower energy consumption of embedded devices. In \cite{schwermer2023energy}, the authors propose the Distributed Edge Device Testbed (DEDT) to evaluate a convolutional neural network trained on the MNIST dataset under various network constraints, quantifying network traffic, energy consumption, and training time. Their results highlight the importance of running FL experiments on physically separated nodes instead of on virtual machines or on one machine with simulated clients \cite{schwermer2023energy}.  There are also several existing solutions that propose energy efficient training strategies to improve the sustainability of Federated Learning \cite{zeng2020energy, guler2021framework, brum2022optimizing, yousefpour2023green}.  However, they mainly focus on reducing the energy cost of model training only for cross-device Federated Learning and lack of sustainability comparison with the centralized solution.

Existing solutions in the literature exhibit several common limitations:
\begin{itemize}
	\item The majority of advanced studies concentrate on the sustainability aspects of cross-device Federated Learning, largely neglecting Cross-Silo Federated Learning. 
    
    \item Existing approaches predominantly assess the energy costs associated with the training phase, without considering the entire lifecycle of AI products.
    
    \item Most investigations are conducted within simulated Federated Learning environments rather than being applied to real-world scenarios and environments.  
\end{itemize}

In this work, we propose to deal with these limitations and provide a more holistic analysis of the carbon footprint of Cross-Silo Federated Learning in the real world environment.

\section{Quantifying CO2 estimations}

\subsection{The real world Federated Learning environment}

\begin{figure}
	    \centering
	    \includegraphics[width=0.5\textwidth]{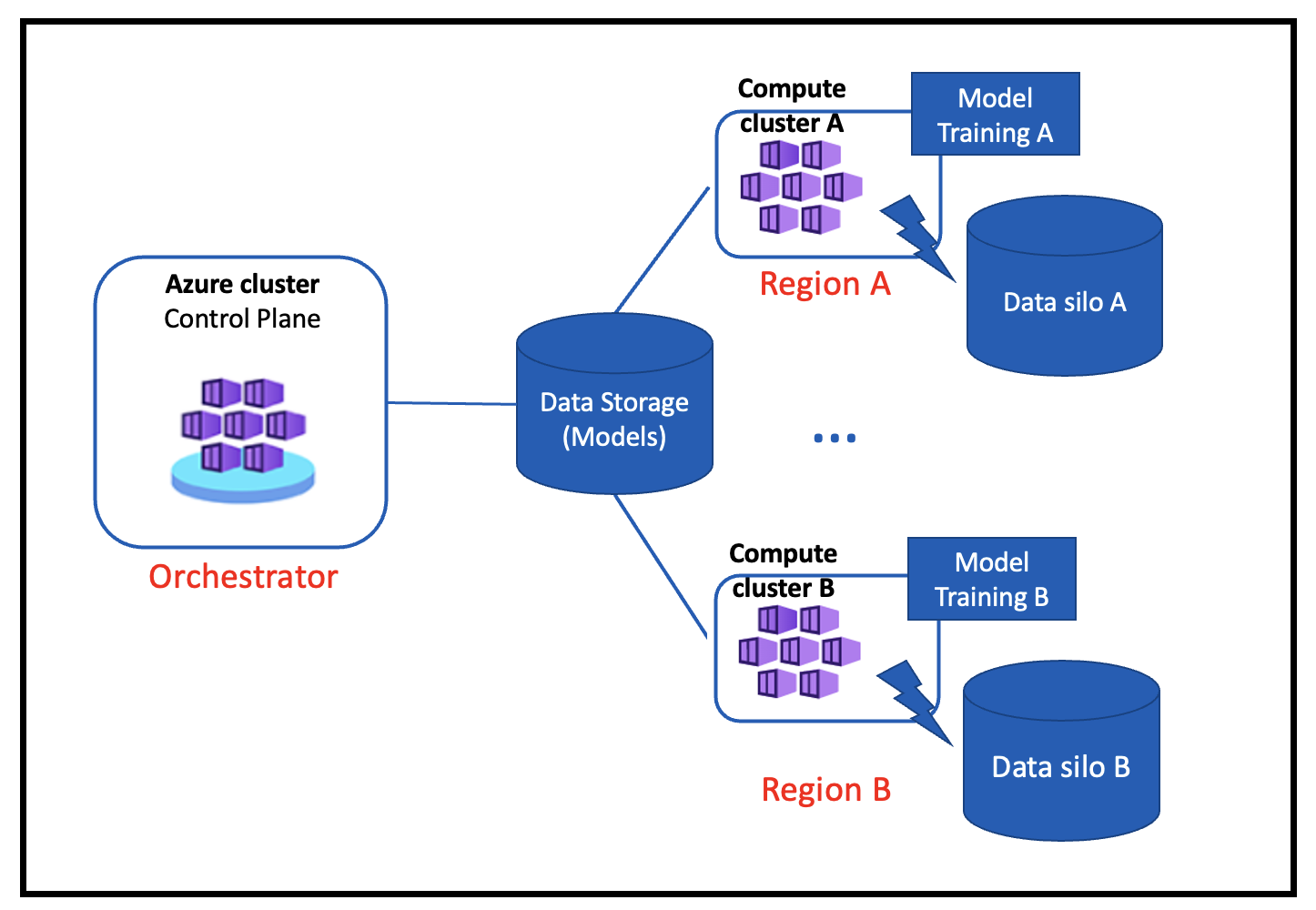}
	    \caption{The real world Cross-Silo Federated Learning setting on the cloud with the example of two silos and one orchestrator.}
	    \label{fig:set}
	\end{figure}

In this work, the real world setting of Cross-Silo Federated Learning environment is constructed on Azure Cloud following the suggestions from \cite{msftfl} as shown in Figure \ref{fig:set}:  it can be seen that each data silo is accessed only by the corresponding compute allocated to this silo for the local training while the Orchestrator has no access to data silos. The role of the Orchestrator (the central server) is to aggregate the weights of local models and update the global model following a strategy such as FedAVG \cite{mcmahan2017communication, yang2019federated}. There is one data storage which is shared by the Orchestrator and local computes to exchange the model weights, where each local compute save the weights to this data storage after local training and the Orchestrator aggregate the local models' weights and save the global model weights also to this storage. 

\subsection{CO2 estimation on the Cloud}

As the real world setting of Cross-Silo Federated Learning environment is on the Cloud, we focus mainly on measuring the CO2 estimation for Cloud providers in this work. The Software Carbon Intensity (SCI), as defined by the Green Software Foundation, is a rate of carbon emissions per one functional unit or R \cite{dodge2022measuring}. The  SCI value measurement equation of a software system is: 
\begin{equation}
    SCI = ((E \times I)+M) \text{  per R} 
    \label{sci}
\end{equation}
 where E is the energy consumption (kilowatt hours) for different components of the software boundary over a given time period, including CPU/GPUs at different percentages of utilisation, data storage, memory allocation, data transferred over a network, etc. I is the Emissions factor measured by grams of carbon dioxide equivalent per kilowatt-hour of electricity (gCO2eq/kWh), which represents the Location-based marginal carbon emissions for the grid that powers the datacenter. M is embodied carbon emission related to the creation, usage, and disposal of a hardware device; R is the functional unit \cite{sci}.

In the lifecycle of an AI project within a company, the initial phase involves identifying pertinent datasets following the clear definition of the problem at hand. In a Centralized Learning environment, this necessitates requesting access to these datasets and subsequently transferring them to the team's designated storage systems. Conversely, Federated Learning eliminates the need for additional data copying, transfer, or centralized storage, thereby streamlining this process. The subsequent phase involves data processing, model training, and testing, where the primary contributors to CO2 emissions are the compute instances or clusters utilized. During this stage, Cross-Silo Federated Learning generally employs a higher number of smaller-size compute instances or clusters in comparison to Centralized Learning. The concluding phase in both methodologies is model deployment and monitoring, which remains consistent irrespective of the learning approach employed. 

\begin{figure*}
	    \centering
	    \includegraphics[width=0.99\textwidth]{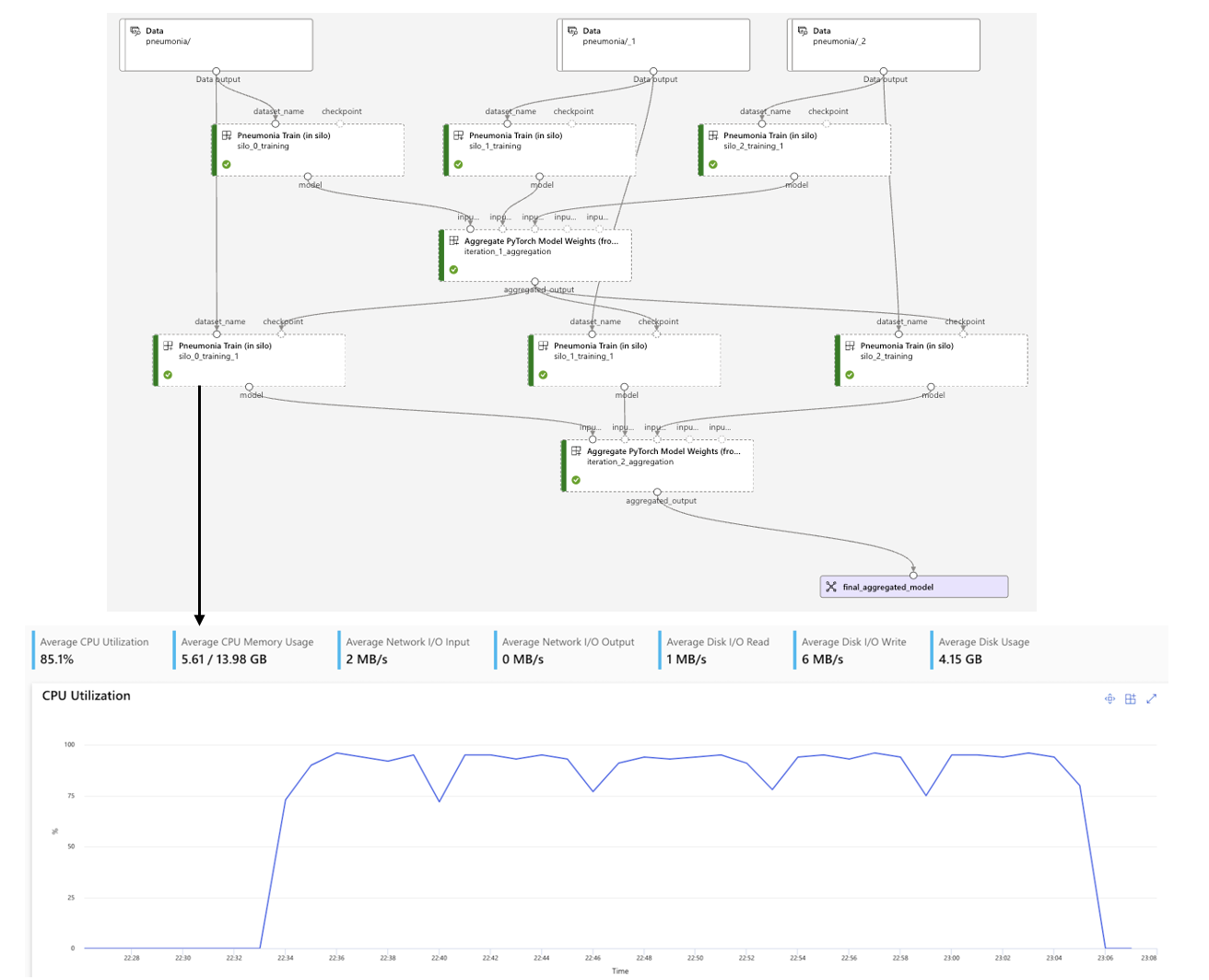}
	    \caption{The example of the real world Cross-Silo Federated Learning training on Azure Cloud.}
	    \label{fig:flex}
	\end{figure*}

Several frameworks exist for measuring CO2 emissions from data centers, including Etsy's Cloud Jewels model, which uses cloud energy conversion factors \cite{etsy}. The Cloud Carbon Footprint (CCF) approach expands on the Cloud Jewels model by estimating CO2e emissions for cloud computing and storage services, and also incorporates networking and memory usage.
Based on the CCF framework, the following metrics are considered in this work:
\begin{itemize}
    \item Storage: 0.65 Watt-Hours per Terabyte-Hour for HDD. 1.2 Watt-Hours per Terabyte-Hour for SSD. 
    \item Networking: CCF takes into account only the data exchanged between different geographical data centers and use the smallest coefficient available to date: 0.001 kWh/Gb for data transfer. However, based on \cite{msft, aslan2018electricity} the average electricity intensity of transmitting data through the Internet (measured as kilowatt-hours per gigabyte [kWh/GB] is: 0.06.
    \item Memory: CCF takes the average of both different estimations and uses 0.000392 Kilowatt Hour / Gigabyte Hour for memory energy consumption. 
    \item Power Usage Effectiveness (PUE): PUE is a score of how energy efficient a data center is, which needs to be multiplied with estimated kilowatt hours for compute, storage and networking. Based on CCF, the PUEs for different Cloud providers are: AWS: 1.135, GCP: 1.1, Azure: 1.185
    \item Carbon Intensity (CI): Convert estimated kilowatt hours for usage of a given cloud provider to estimated CO2e based on the cloud provider data center region that each service is running in. Google has published the grid carbon intensity for their GCP regions. For AWS and Azure, CCF  generally uses carbonfootprint.com’s country specific grid emissions factors report \cite{ccf}. For most of Europe, EEA emissions factors are used \cite{eea}. 

\end{itemize}

Cloud providers typically offer storage, database services, and node pools that inherently replicate data along with associated computational and memory resources to ensure high data durability and availability. This built-in redundancy is crucial for maintaining service continuity during outages.
For example,\textit{ "Azure Storage always stores multiple copies of your data so that it's protected from planned and unplanned events, including transient hardware failures, network or power outages, and massive natural disasters"}\footnote{https://learn.microsoft.com/en-us/azure/storage/common/storage-redundancy}. 
Locally redundant storage (LRS) achieves this by synchronously replicating your data three times within a single physical location in the primary region. On the other hand, zone-redundant storage (ZRS) synchronously replicates data across three different Azure availability zones within the primary region. It is essential to account for data redundancy when evaluating the carbon footprint associated with data transfer and storage.

To estimate the energy consumption of computing components like CPUs and GPUs, we employ the methodology illustrated by Microsoft in their study \cite{compute}. This approach utilizes the Thermal Design Power (TDP) values of these hardware elements to approximate their power usage, as detailed in Equation \ref{ec}: 

\begin{equation}
    E_c = N_c \times TDP \times load \times T /1000 \text{ (kWh)} 
    \label{ec}
\end{equation}
where $E_c$ is the energy consumption of computing, $N_c$ is the number of computes such as CPUs and GPUs, $load$ is the compute utilization (as shown in the example in Figure \ref{fig:flex}), $T$ is the computing time (hours). 

Once we get $E_c$, the CO2 emission estimation caused by computing is calculated as:
\begin{equation}
    C_c = E_c \times PUE \times CI
    \label{cc}
\end{equation}
where PUE is the Power Usage Effectiveness defined above, CI is the region specific Carbon Intensity defined above. The equations to estimate the CO2e of memory, data transfer and data storage are omitted as they are similar to Equation \ref{ec} and  Equation \ref{cc}.

\section{Experiments}

\begin{figure}
	    \centering
	    \includegraphics[width=0.5\textwidth]{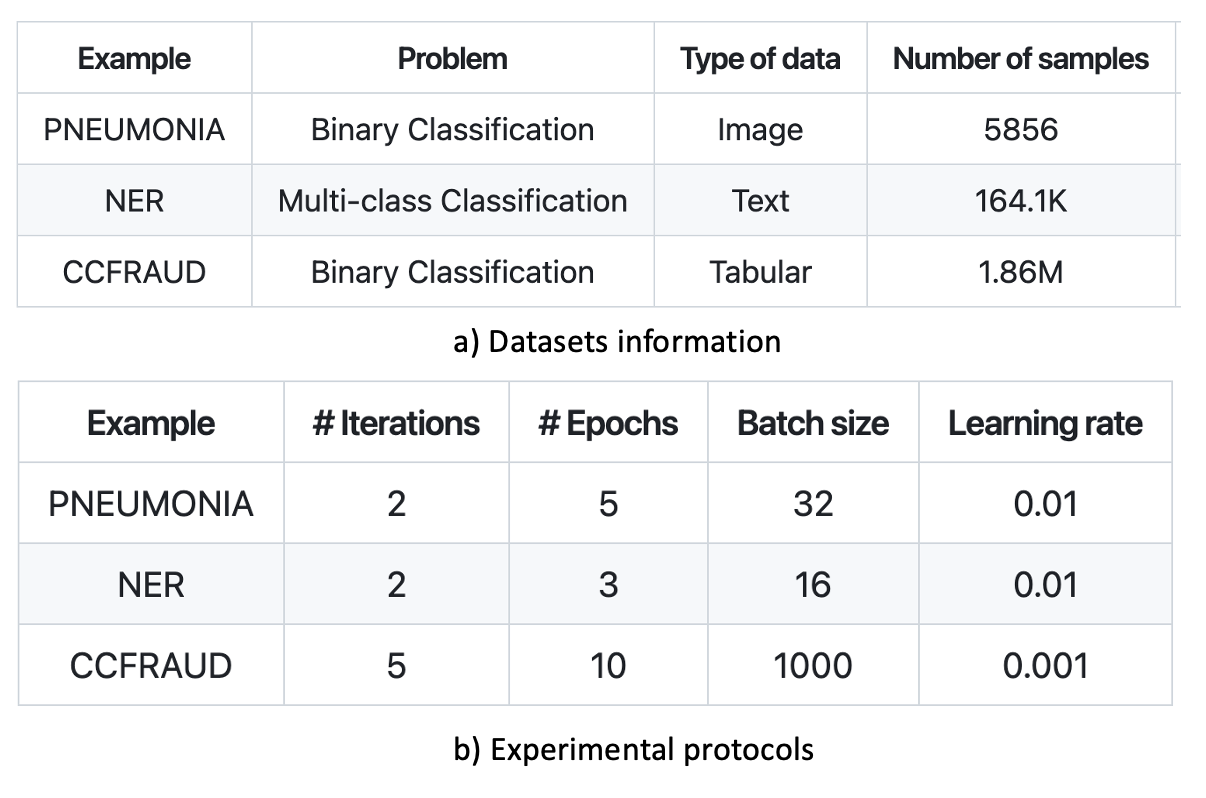}
	    \caption{The datasets and experimental protocols used in this study.}
	    \label{fig:data}
	\end{figure}

This research aims to conduct a detailed analysis of the CO2 emissions associated with Cross-Silo Federated Learning as opposed to Centralized Learning within a real-world environmental setting. Given that both methodologies exhibit comparable levels of embodied emissions and similar CO2 emissions during model deployment and monitoring phases, our primary focus is directed towards evaluating the differences in CO2 emissions arising from data storage, data transfer, network, and model training processes.

\subsection{Datasets and experimental protocols}

In this study, we adopt the datasets and experimental protocols outlined in Microsoft's benchmarking study, which are detailed in their publicly available documentation  (as shown in Figure \ref{fig:data}) \footnote{https://github.com/Azure-Samples/azure-ml-federated-learning/blob/main/docs/concepts/benchmarking.md}.  This benchmark evaluated the training overhead, model performance, and scalability of Federated Learning in a Cross-Silo configuration on the Azure Cloud. Their results demonstrated that Federated Learning across multiple silos enhances model performance compared to training a single model with partial datasets, and that its performance is on par with Centralized Learning, where a single model is trained on the entire dataset. Building on this foundation, our research specifically aims to compare the CO2 emissions and cost implications between Cross-Silo Federated Learning and Centralized Learning, utilizing the same data, models, and experimental protocols.

To assess the scalability of our approach, we replicated the datasets in a manner consistent with the Microsoft benchmark. We then evaluated the performance of Cross-Silo Federated Learning and Centralized Learning across three dataset scales: a small dataset of approximately 1.2 GB, a medium dataset of around 12 GB, and a large dataset of roughly 120 GB. These dataset sizes are representative of typical Machine Learning applications within corporate environments.

\subsection{Experimental results}

\begin{figure}
	    \centering
	    \includegraphics[width=0.5\textwidth]{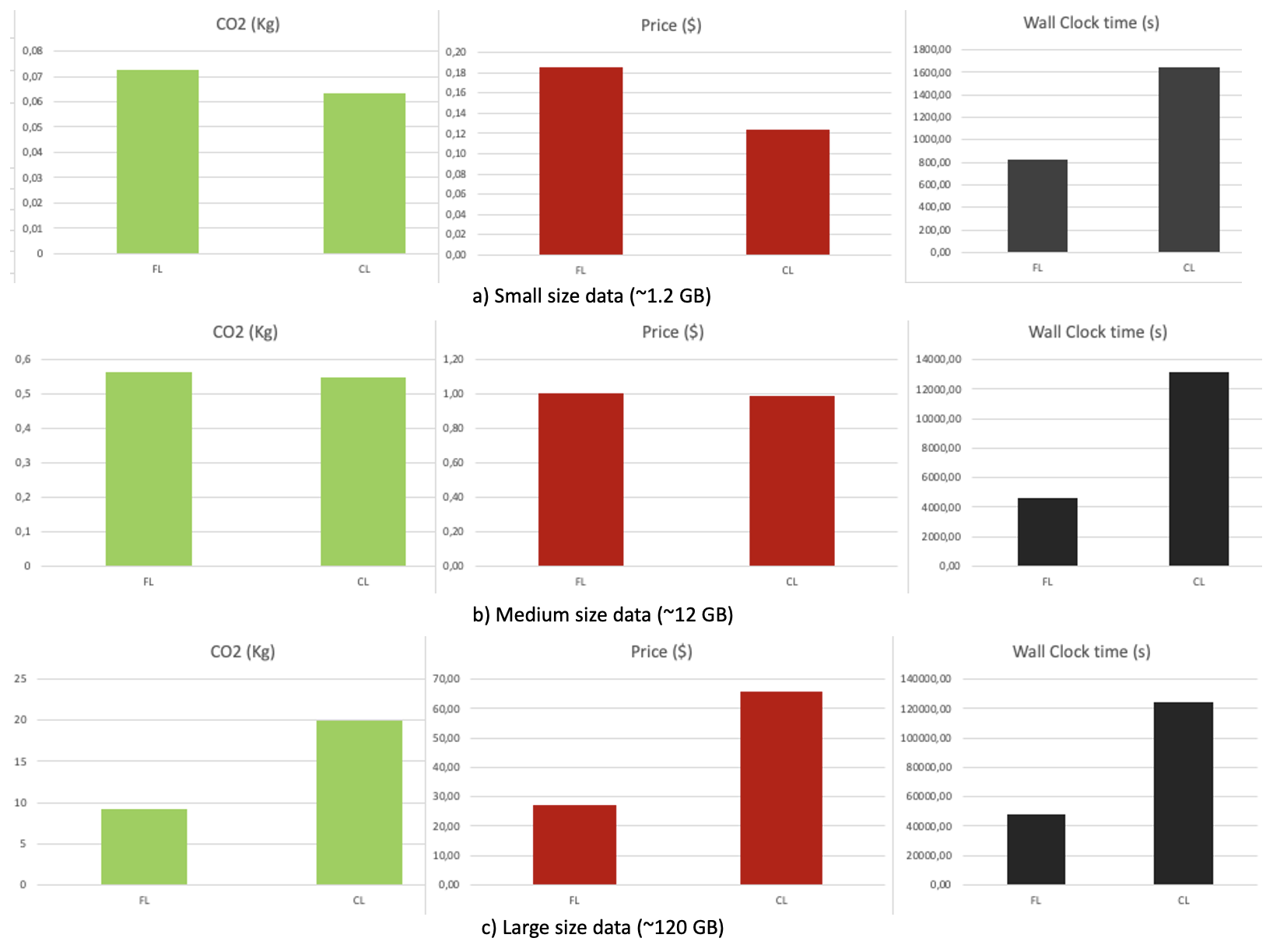}
	    \caption{The comparison between Cross-Silo Federated Learning and Centralized Learning in terms of CO2 emission, cost and computing time during model training.}
	    \label{fig:train}
	\end{figure}

Firstly, the comparison between Cross-Silo Federated Learning and Centralized Learning in terms of model training is shown in Equation \ref{ctrain}: including CO2 emissions of CPU/GPU utilization ($C_{CPU}$, $C_{GPU}$), memory consumption ($C_{memory}$), and additional communication overhead ($C_{network}$) inherent to FL.  

\begin{equation}
    C_{train} = C_{CPU} + C_{GPU} + C_{memory} + C_{network}
    \label{ctrain}
\end{equation}

The real-world experimental results across small, medium and large data sizes are illustrated in Figure \ref{fig:train}, comparing CO2 emissions, costs, and computing times between Cross-Silo Federated Learning and Centralized Learning  during model training phase. The results indicate that distributed training in Cross-Silo FL significantly reduces wall clock time across various dataset sizes. For small to medium datasets, Cross-Silo FL incurs slightly higher CO2 emissions and costs compared to CL due to the additional communication rounds required. However, for large datasets, CL results in greater CO2 emissions and costs relative to Cross-Silo FL. This is because CL necessitates the use of larger compute clusters to handle the increased data volume. It is important to note that a basic compute cluster selection method based on dataset size was used in this work, presenting an opportunity for further optimization.

\begin{figure}
	    \centering
	    \includegraphics[width=0.5\textwidth]{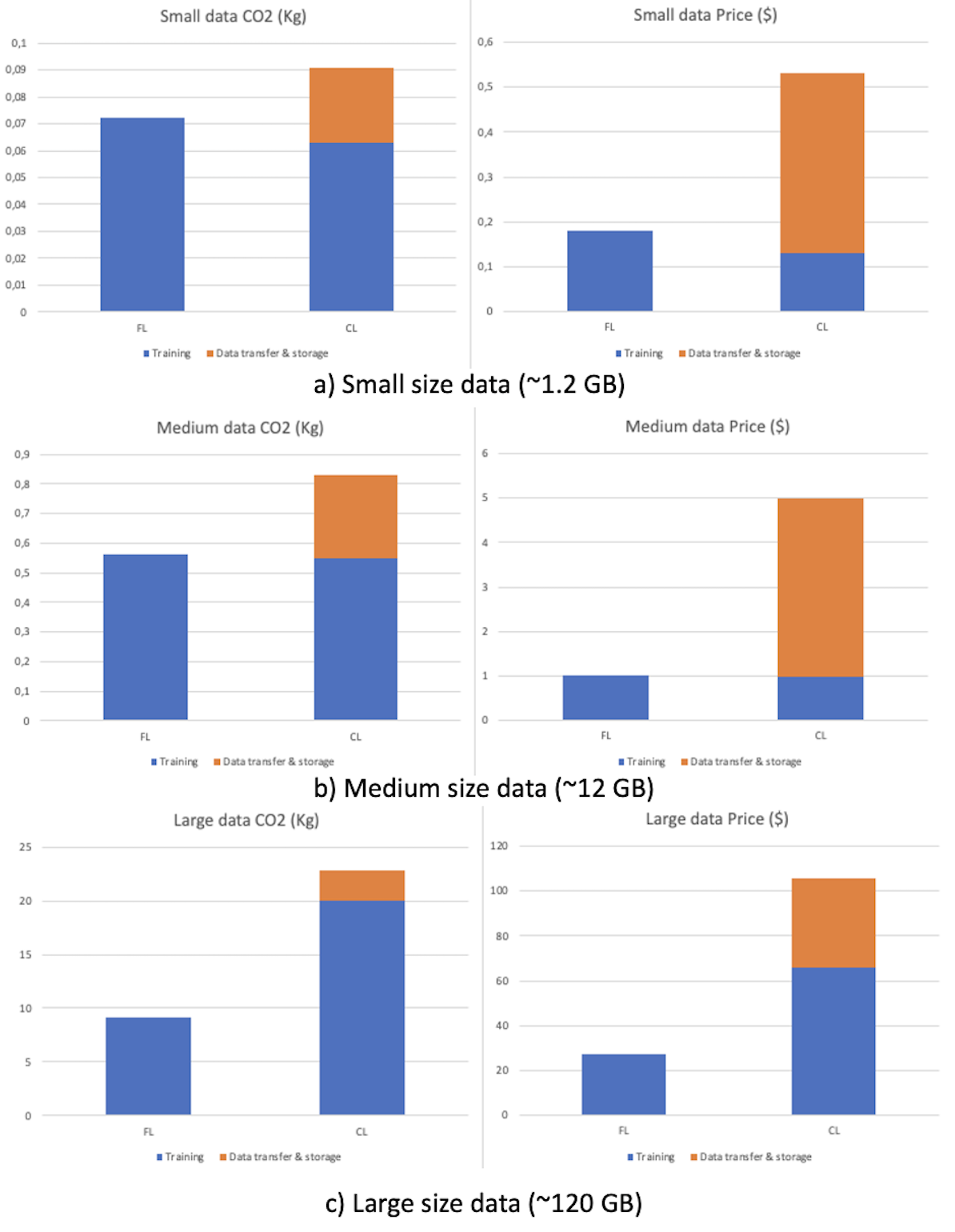}
	    \caption{The comparison between Cross-Silo Federated Learning and Centralized Learning in terms of CO2 emission and cost including model training, data transfer and data storage.}
	    \label{fig:all}
	\end{figure}
    
Secondly, the comparison between Cross-Silo Federated Learning and Centralized Learning across the AI product lifecycle is shown in Equation \ref{ctotal}:  including model training ($C_{train}$), data transfer ($C_{transfer}$: extra  cost of Centralized Learning) and data copy storage ($C_{storage}$: extra  cost of Centralized Learning).  

\begin{equation}
    C_{total} = C_{train}  + C_{transfer} + C_{storage}
    \label{ctotal}
\end{equation}

The real-world experimental results are illustrated  in Figure \ref{fig:all}. 
The blue bars indicate the CO2 emissions and costs associated with model training, while the orange bars represent the additional CO2 emissions and costs incurred from data storage and transfer. 
The findings reveal a divergence from the results typically reported in the literature on Federated Learning, which usually consider model training in isolation. When accounting for both data storage and transfer alongside model training, Centralized Learning consistently exhibits higher CO2 emissions and greater overall costs compared to Cross-Silo Federated Learning, regardless of the training dataset size.

In summary, this study provides a comprehensive comparison between Cross-Silo Federated Learning  and Centralized Learning  with respect to cost and CO2 emissions. It evaluates these learning paradigms by examining data transfer, data storage, CPU/GPU usage, memory usage, networking, and the additional communication necessary for FL. The findings reveal that Cross-Silo FL achieves significantly lower wall clock times across various dataset sizes, attributable to the distributed nature of the training process. Although the energy consumption and costs associated with model training are comparable between Cross-Silo FL and CL, the centralized approach incurs more CO2 emissions due to increased data transfer and storage needs. These factors are frequently overlooked in the state-of-the-art analysis, highlighting the importance of considering them in future evaluations.

\section{Proposed Data and application management system}

\begin{figure*}[h]
	    \centering
	    \includegraphics[width=0.89\textwidth]{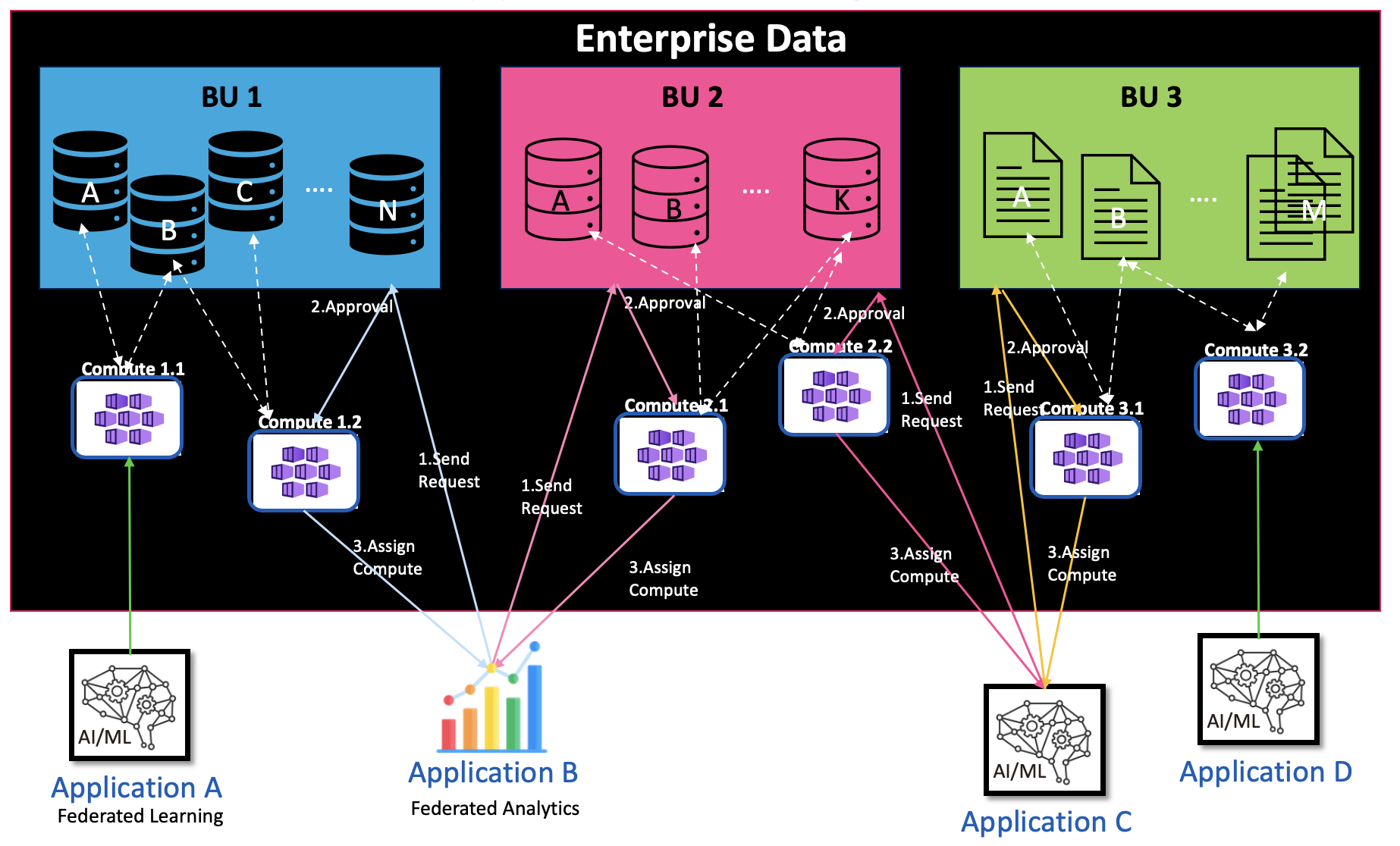}
	    \caption{The proposed Sustainable Data and Application management system leveraging Cross-Silo Federated Learning and Federated Analytics: Application owners submit use cases to data owners, which are checked for redundancy using historical requests and LLMs. Approved requests are granted data access and compute resources. FL and FA then enable privacy-preserving, resource-efficient development across organizational data silos.}
	    \label{fig:propose}
	\end{figure*}

Building on the experimental results and analyses presented in the previous section, this section explores the potential of utilizing Cross-Silo Federated Learning in conjunction with Federated Analytics to minimize data redundancy, model redundancy, and application redundancy within organizations. This approach aims to enhance data privacy, streamline business operations, and promote cost-effectiveness and environmental sustainability in corporate solutions. 

The architecture of the proposed data and application management system is illustrated in Figure \ref{fig:propose}. In today's technological landscape, data are invaluable assets for companies. Consider a scenario where a company possesses multiple data silos, each managed by different business units across various regions. To enhance the sustainability of enterprise data and application management for each application or product, the following steps can be taken instead of duplicating data requests:
\begin{itemize}
    \item The application or product owner should clearly outline the use case in a detailed text description and submit this request to the data owners.
    \item Data owners will then evaluate the request and compare it with historical requests, potentially utilizing Large Language Models for this analysis. If the application or product is new, the request will be approved, and an appropriate compute cluster will be assigned to the application or product owner, granting read access to the requested data. This approach minimizes resource wastage by ensuring the correct compute cluster is selected. If the application or product already exists, the contact information of the existing application or product owners will be shared to prevent redundancy.
    \item The application or product owner will then employ Cross-Silo Federated Learning or Federated Analytics for further development (solutions in the multi-view learning literature such as \cite{cao2018dynamic, cao2021novel, cao2019randomphd, hongliu2024user} can be used too).
This method ensures efficient use of data and computational resources, fostering a more sustainable and collaborative environment for data and application management within the enterprise.
\end{itemize}

\section{Conclusion}

This study introduces a novel framework for evaluating the energy and carbon footprints associated with Cross-Silo Federated Learning throughout the AI product lifecycle, moving beyond the traditional focus on model training found in existing literature. This research is the first to take an enterprise data and product management perspective, comparing Cross-Silo Federated Learning with centralized methods. Our experiments, conducted in a real-world Federated Learning environment on Azure Cloud, aim to assist companies in becoming more cost-effective and environmentally sustainable. 
The key findings from our experiments indicate that the energy consumption and costs of model training are comparable between Cross-Silo Federated Learning and Centralized Learning. However, the additional data transfer and storage requirements inherent in Centralized Learning can result in significant, often overlooked CO2 emissions. 
Based on our experimental analysis, we propose a Cross-Silo Federated Learning and Analytics framework designed to help companies minimize data, model, and application redundancy. This approach not only enhances privacy but also contributes to reducing the environmental impact across AI products' lifecycle.

\iffalse
This research also has a few  limitations. Firstly, the comparison of performance between Cross-Silo Federated Learning and Centralized Learning is not exhaustive. The benchmark datasets utilized in this study are predominantly Independent and Identically Distributed (IID). Future work should also evaluate non-IID datasets in the context of Federated Learning. Moreover, this study does not encompass real-world experiments involving Federated Analytics, which focuses on collaborative data analysis without model training. Future investigations should also address this aspect.
\fi

\bibliography{sample}

\end{document}